%% file: main.tex
\documentclass[letterpaper, 10 pt, conference]{ieeeconf}  

\IEEEoverridecommandlockouts                              

\usepackage{graphicx} 
\usepackage{epsfig} 
\usepackage{times} 
\usepackage{amsmath} 
\usepackage{amssymb}  
\usepackage{color}
\usepackage{booktabs}
\usepackage[caption=false]{subfig}

\title{\LARGE \bf
Distant Vehicle Detection Using Radar and Vision 
}

\author{Simon Chadwick, Will Maddern and Paul Newman
\thanks{Authors are from the Oxford Robotics Institute, Dept. Engineering Science, University of Oxford, UK.
\texttt{\{simonc,wm,pnewman\}@robots.ox.ac.uk}}}

\begin{document}

\maketitle
\thispagestyle{empty}
\pagestyle{empty}

\begin{abstract}

For autonomous vehicles to be able to operate successfully they need to be aware of other vehicles with sufficient time to make safe, stable plans. Given the possible closing speeds between two vehicles, this necessitates the ability to accurately detect distant vehicles. Many current image-based object detectors using convolutional neural networks exhibit excellent performance on existing datasets such as KITTI. However, the performance of these networks falls when detecting small (distant) objects. We demonstrate that incorporating radar data can boost performance in these difficult situations. We also introduce an efficient automated method for training data generation using cameras of different focal lengths. 

\end{abstract}

\section{Introduction}

In an autonomous driving setting, object detection, particularly the detection of other road users, is of vital importance for safe operation. However, in some circumstances the closing speeds between vehicles can be such that it becomes necessary to detect objects at considerable distances to be able to make timely decisions. At such distances, the appearance of a vehicle in an image is only a few pixels high and wide.

Deep convolutional neural networks (CNNs) represent the state-of-the-art in object detection and have been shown to perform well in a range of different scenarios. However, it has also been shown that they struggle to accurately detect small objects \cite{liu2016ssd}.

Radar as a sensing modality is highly complementary to vision. It is  robust to variable weather conditions, has measurement accuracy that is independent of range and, in the case of Doppler radar, provides direct velocity measurements. However, unlike images it is difficult to determine the size of a radar target \cite{stanislas2015characterisation} and radar signals are vulnerable to clutter and multi-path effects. We show that the combination of the two sensors improves performance over vision alone. In particular, Doppler radar can give a strong indication of motion for objects in the distance helping to classify small groups of pixels as, for example, a vehicle rather than as background.

We design and train a detector that operates using both monocular images and radar scans to perform robust vehicle detection in a variety of settings. None of the major datasets used for autonomous driving research \cite{Geiger2012CVPR}\cite{Cordts2016Cityscapes}\cite{RobotCarDatasetIJRR} include radar data. As the performance of CNNs is closely linked to the availability of sufficient training data it is important that there is a low cost source of this data. Consequently, to demonstrate our approach, we generate our own dataset that we label automatically by using an existing detector and combining detections from multiple cameras. By using cameras of different focal lengths we can label distant objects and exploit the proximity of the cameras to transfer the distant labels into the canonical image. 

\begin{figure}[t!]
    \centering
    \includegraphics[width=\columnwidth]{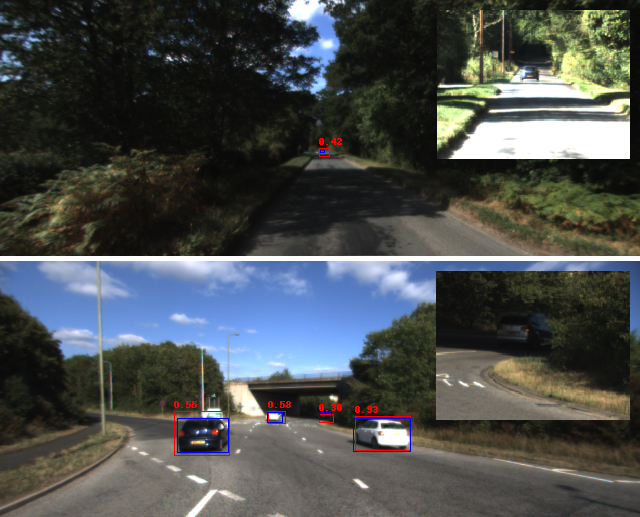}
    \caption{By using radar we are able to detect vehicles even if they are very small (\textit{top}) or hard to see (\textit{bottom}). The inset images show the difficult parts of the main scenes and are taken from a synchronised long focal length camera we use as part of our training data generation. Detections are shown in red, ground truth in blue. This figure is best viewed electronically.}
    \label{fig:placeholder}
\end{figure}
   
\section{Related Work}

Object detection has been a major topic of computer vision research and over recent years a number of fully convolutional object detectors have been proposed. Two-stage methods, in particular Faster R-CNN \cite{ren2015faster}, provide state-of-the-art performance but are computationally expensive. One-stage methods \cite{liu2016ssd}\cite{redmon2016you} are structurally simpler and can operate in real-time but suffer a performance penalty. To improve performance on difficult examples and bridge the performance gap between two-stage and one-stage detectors, Lin \textit{et al.} \cite{lin2017focal} propose a loss function that focuses the loss on examples about which the classifier is least confident. This however relies on the labels being highly accurate which may not be the case if they are automatically generated.

A common approach when training object detectors for a specific task is to pre-train a feature extractor using ImageNet \cite{russakovsky2015imagenet} and then fine tune the features with the limited training data available for the task. In \cite{shen2017dsod} Shen \textit{et al.} show that, given careful network design, it is possible to obtain state of the art results without this pre-training process. This implies that a fusion network, such as the one we are proposing, is not at an insurmountable disadvantage if pre-training is not performed. 

A number of papers use automated methods for generating training labels. In \cite{barnes2017find}, visual odometry from previous traversals is used to label driveable surfaces for semantic segmentation. Hoermann \cite{hoermann2018object} employs temporal consistency to generate labels by processing data both forwards and backwards in time. Recent work by Adhikari \textit{et al.} \cite{adhikari2018faster} takes a labelling approach that is related to ours by also leveraging the power of an existing object detector to generate labelled data for a new task, although a small amount of manual labelling is still required.

Multi-modal object detection has been investigated in a number of works. In \cite{chen2017multi} camera images are combined with both frontal and birds eye LIDAR views for 3D object detection. Data from cameras, LIDAR and radar are all fused in \cite{chavez2016multiple}.

\section{Dataset Creation and Automatic Labelling }

\begin{figure}
\vspace{1em}
\centering
\includegraphics[width=\columnwidth]{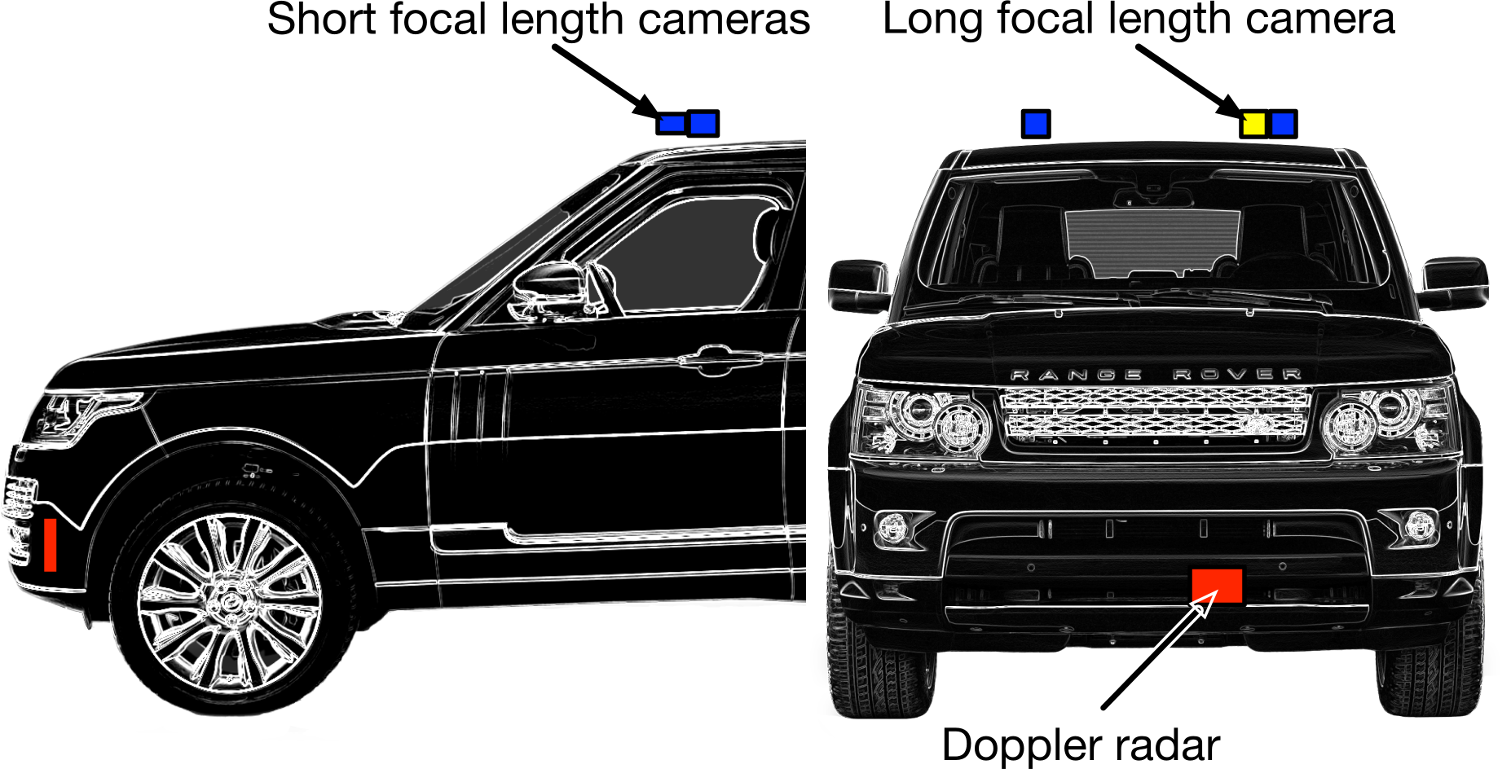} 
\caption{The sensor layout used for data collection. Short focal length (wide angle) cameras are shown in blue, long focal length in yellow and radar in red.}
\label{fig:sensor_layout}
\vspace{-1em}
\end{figure}

To create our dataset, data is gathered using two cameras configured as a stereo pair and a third, with a long focal length lens, positioned next to the left stereo camera (Fig. \ref{fig:sensor_layout}). All three cameras are synchronised and collect 1280x960 RGB images at 30Hz. In addition we collect radar data using a Delphi ESR 2.5 pulse Doppler cruise control radar with a scan frequency of 20Hz. The radar is dual-beam, simultaneously operating a wide angle medium range beam (\(>90\deg\), \(>50m\)) and a long range forward-facing narrow beam (\(>20\deg\), \(>100m\)). Labels are generated in an automated fashion --- the way in which we do this is part of our contribution. 

\subsection{Object detector}\label{object_detector}

To generate our labels we use an implementation of the YOLO object detector \cite{redmon2016you} trained on the KITTI dataset \cite{Geiger2012CVPR}. This detector provides a variable number of bounding box detections per frame with each detection having an object probability \(P(O)\) and conditional class probabilities \(P(C|O)\) in addition to bounding box co-ordinates. However, the unmodified object detections from a single image are insufficiently accurate to use as labels, particularly for objects at distance. To improve the quality of the detections we combine detections from multiple cameras. In the implementation we use, two classes are provided: vehicles and pedestrians/cyclists.

\subsection{Combining labels using multiple focal lengths}\label{focal_lengths}

\begin{figure*}[t!]
    \vspace{1em}
    \centering
    \includegraphics[width=\textwidth]{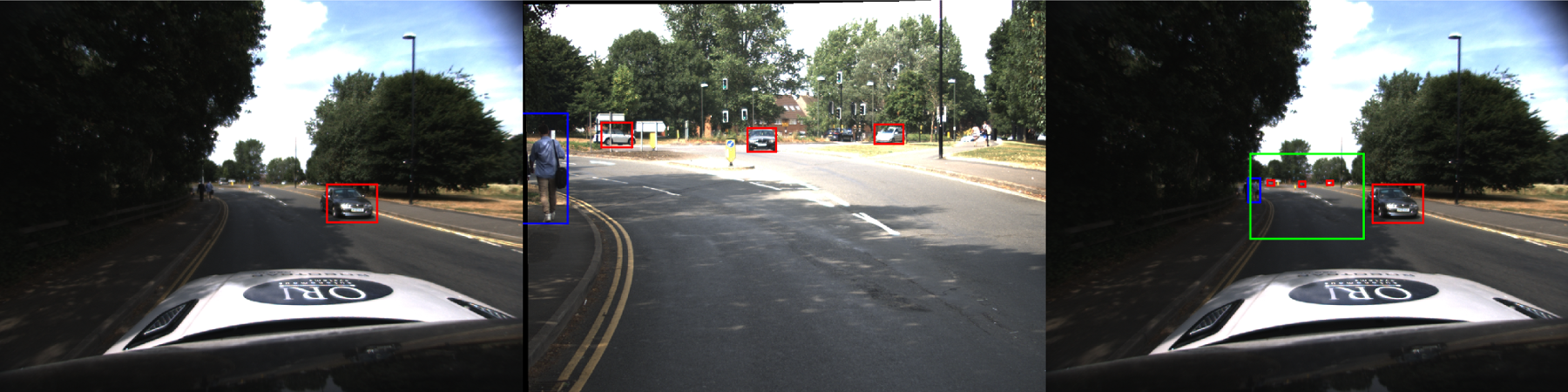}
    \caption{Example of bounding box transfer between two cameras of different focal lengths for training data generation. \textit{Left} shows the original bounding boxes found from the short focal length camera (vehicles are red, pedestrians blue). \textit{Middle} shows the original bounding boxes found from the long focal length camera. \textit{Right} shows the combined set of bounding boxes. The outline of the overlapping region is shown in green.}
    \label{fig:multi_camera_boxes}
    \vspace{-1em}
\end{figure*}

To produce more accurate labels of distant vehicles we make use of two cameras of different focal lengths. The first camera \(C_{A}\) has a wide angle lens (short focal length) and is the camera in which objects are to be detected when the system is deployed live on a vehicle. The second camera \(C_{B}\) has a much longer focal length and is mounted as close as physically possible to the first such that their optical axes are approximately aligned. Object detections in \(C_{B}\) can be transferred to \(C_{A}\) without needing to know the object's range (Fig. \ref{fig:multi_camera_boxes}) by exploiting the cameras' close mounting. Specifically, given two cameras, \(C_{A}\) and \(C_{B}\), with respective intrinsic matrices \(\mathbf{K}_{A}\) and \(\mathbf{K}_{B}\), if the cameras are mounted close together, such that the camera centres can be approximated as being coincident, then image points \([u, v]\) from one camera can be redrawn in the other camera as follows: 
\begin{equation}
    \mathbf{x} = 
        \begin{bmatrix}
            uw\\
            vw\\
            w
        \end{bmatrix}
\end{equation}
\begin{equation}
    \mathbf{x}_{A} = \mathbf{K}_{A}\mathbf{R}_{AB}\mathbf{K}_{B}^{-1}\mathbf{x}_{B}
\end{equation}
\noindent where \(\mathbf{x}_{A}\) and \(\mathbf{x}_{B}\) are the homogeneous image co-ordinates in the respective cameras, \(\mathbf{R}_{AB}\) is the rotation matrix between the two cameras and \(w\) is set to \(1\). \(\mathbf{R}_{AB}\) is obtained by minimising the re-projection error of salient points that are manually identified in synchronised images from both cameras. 

The approximation that both cameras share a camera centre induces an error \(\epsilon\) in the points that are transferred to the new image:
\begin{equation}
    \epsilon = \frac{f_{A}d}{Z}
\end{equation}
where \(d\) is the distance between the two cameras, \(f_{A}\) is the focal length of the new camera and \(Z\) is the distance of the point from the camera along the optical axis. In our application with \(f_{A} = 625px\) and \(d = 0.032\) a bounding box for an object at \(Z = 20m\) away will be offset by \(\epsilon = 1px\).

Given that the fields of view of the cameras are required to overlap as much as possible there will be a joint image region observed by both cameras. It is likely that objects in this joint region will be detected in both the short and long focal length images simultaneously. As the long focal length camera \(C_{B}\) has a higher resolution in the joint region we use the detections from that camera in that region. At the border we discard any detections from the short focal length camera \(C_{A}\) which have an overlap with the joint region greater than a threshold \(\tau\) (we use \(\tau = 0.5 \)). Given a bounding box in \(C_{B}\) with area \(\Phi_{B}\) and a joint region between the two cameras with area \(\Phi_{J}\) the overlap is calculated as:
\begin{equation}
    Overlap = \frac{\Phi_{B} \cap \Phi_{J}}{min(\Phi_{B}, \Phi_{J})}.
\end{equation}

Advantageously, in our automotive application, distant objects are often in the centre of the image which coincides with the overlapped image region. As a result the number of distant objects that are missed is relatively small compared to the ratio of the two image areas. 

\subsection{Radar}

The radar internally performs target identification from the radar scans and outputs a set of identified targets (access to the raw data is not available). Each target comprises measurements of range, bearing, range rate (radial velocity) and amplitude. 
Each radar scan contains a maximum of 64 targets from each of the two beams. To handle the varying number of targets we project the radar targets into camera \(C_{A}\) giving two extra image channels --- range and range-rate, see Fig \ref{fig:example_data}. We mark each target position in the image as a small circle rather than a single pixel as this both increases the influence of each point in the training process and reflects to some extent the uncertainty of the radar measurements in both bearing and height.

To simplify the learning process, before performing the projection we subtract the ego-motion of the platform from the range rate measurement of each target. To calculate the ego-motion we use a conventional stereo visual odometry system \cite{churchillvo}. As the radar is not synchronised with the cameras we take the closest ego-motion estimate to each radar scan.

The projected image locations are calculated as

\begin{equation}
    \mathbf{x}_{A} = \mathbf{P}_{A}\mathbf{T}_{AR}\mathbf{X}_{R}
\end{equation}

\noindent where \(\mathbf{P}_{A}\) is the 3x4 projective camera matrix of \(C_{A}\), \(\mathbf{T}_{AR}\) is the extrinsic calibration matrix from the radar to the camera and \(\mathbf{X}_{R}\) is the 4x1 homogeneous vector of the radar target.

The range-rate channel is scaled such that a range-rate of 0 corresponds to a pixel value of 127. 

As can be seen in Fig. \ref{fig:example_data} the radar is  sparse and can be inconsistent, there is no guarantee that a moving vehicle will be detected as a target. It is also noisy --- occasional high range-rate targets will briefly appear without any apparent relation to the environment. Neverthless, there is sufficient information in the signal that it can provide a useful guide to vehicle location.

\subsection{Sub-sampling}

As consecutive image frames are highly correlated, there is a diminishing return in including all frames. We select only those frames for which radar and image timestamps are within a small time offset (we choose 10ms) and then sub-sample by a factor of five giving an overall sub-sampling of approximately 1:10 from the original radar frequency.

\subsection{Final dataset}

\begin{figure}
\vspace{1em}
\centering
\includegraphics[width=\columnwidth]{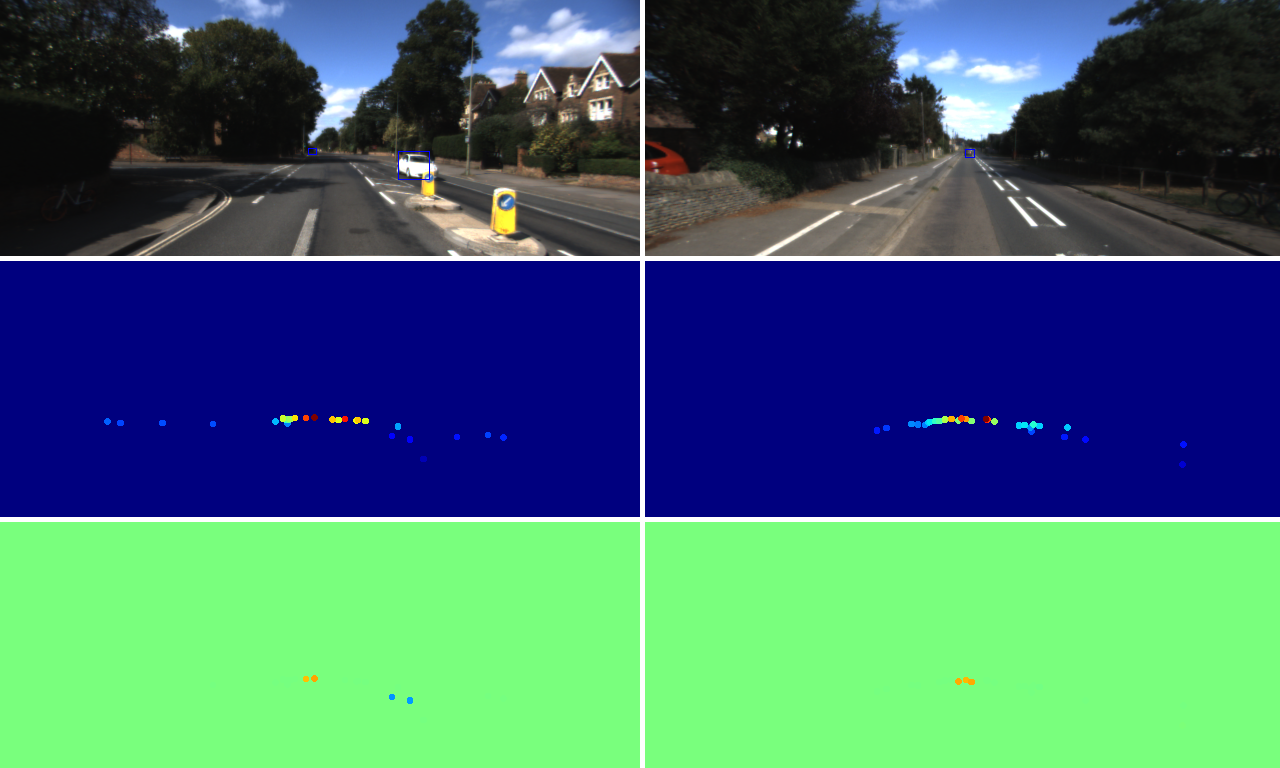} 
\caption{Examples of automatically generated training data. \textit{Top} shows the image with bounding boxes from the object detections from the combined cameras. \textit{Middle} shows the range image generated from the radar scan and \textit{bottom} shows the range-rate image. Note the difference in colour of the signals associated with the oncoming and retreating vehicles in the left-hand range-rate image. The colouring of the radar images is for visualisation only.}
\label{fig:example_data}
\end{figure}

The outcome of the labelling process is a dataset of images and radar scans with bounding boxes for vehicles and pedestrians in each image. From six daytime drives totalling 3 hours 10 minutes that cover a mixture of urban, sub-urban and highway driving, we generate 25076 labelled images. We split the dataset to give 17553, 2508 and 5015 images for training, validation and testing respectively. The images are not shuffled prior to splitting to prevent consecutive (and potentially very similar) images from being split into the training and testing sets and to ensure that unseen locations are present in testing. The images are cropped to remove the vehicle bonnet and then downscaled by a factor of two to give a final image size of 640x256. Some examples from the dataset are shown in Fig. \ref{fig:example_data}. 

Due to a significant class imbalance, with 53019 vehicles in the training set and only 5559 pedestrians, and given that the radar signal on pedestrians is poor, we focus on vehicles only in the rest of this work.

\section{Detection Network}

We build upon the SSD object detection framework \cite{liu2016ssd}, chosen as it represents a proven baseline for single-stage detection networks. We construct our network from ResNet blocks \cite{he2016identity} using the 18-layer variant from \cite{he2016deep}. Using blocks from the larger ResNet variants added model complexity without increasing performance, possibly due to the limited number of classes and training examples (relative to ImageNet) meaning that larger models merely added redundant parameters.

\begin{figure}
\vspace{1em}
\centering
\includegraphics[width=\columnwidth]{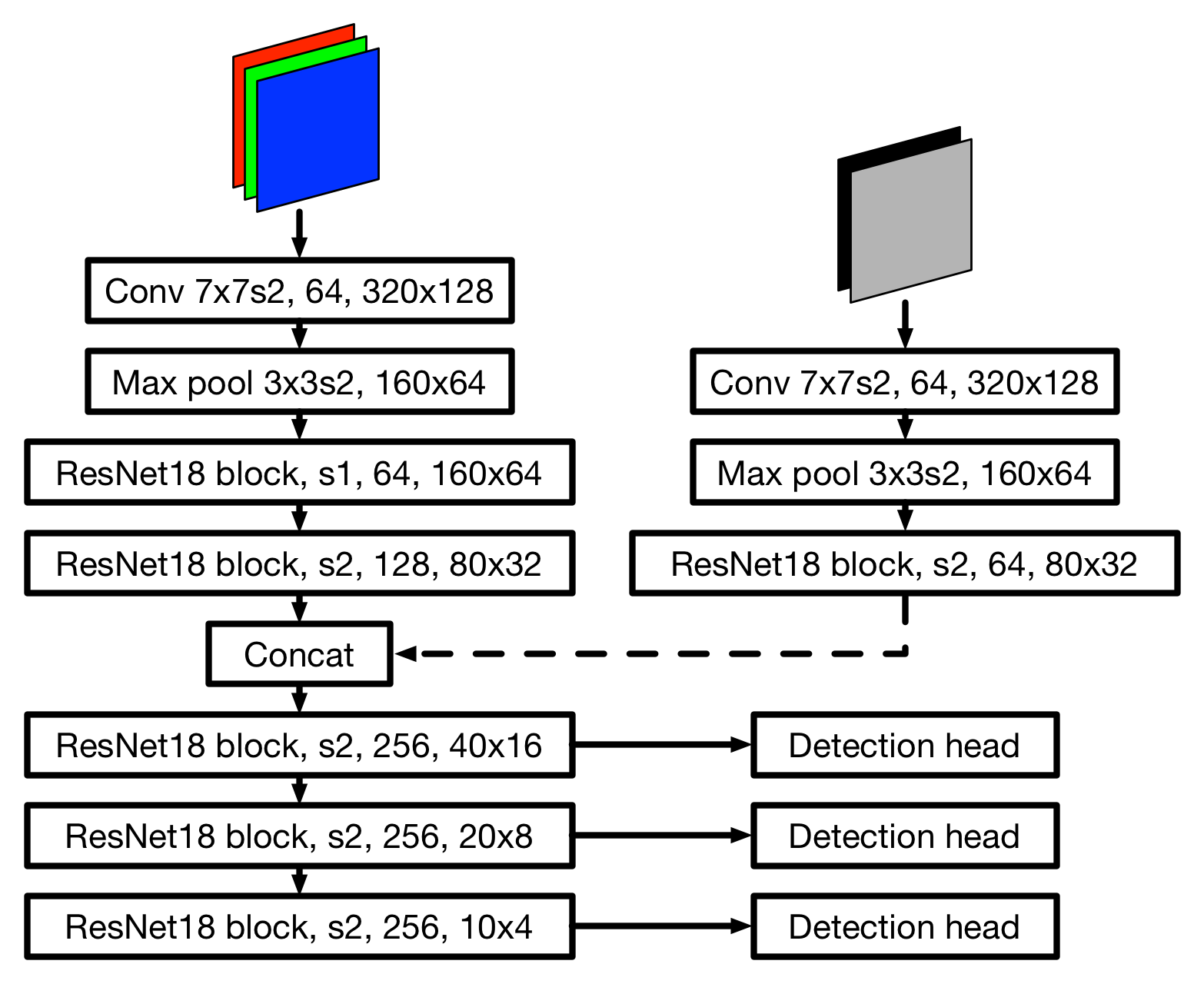} 
\caption{The network configuration for the concatenation fusion version showing filter sizes, strides, output channels and image size for each level. For networks using only RGB images the right-hand radar branch is removed.}
\label{fig:network_layout}
\end{figure}

A variety of possible fusion methods are discussed in recent work \cite{chen2017multi}\cite{jaritz2018sparse}. We experiment with including the radar data in two ways. Firstly, by adding an additional branch for the radar input and concatenating the features after the second image ResNet block (see Fig. \ref{fig:network_layout}). Secondly, by adding the same additional branch but without the max-pool and using element-wise addition to fuse the features after the first image ResNet block. While we experimented with a combined five-channel input image, the branch configuration proved best, allowing the development of separate radar and RGB features. Using a branch structure also offers the potential flexibility of re-using weights from the RGB branch with different radar representations. 

As in standard SSD, features at different scales are passed through classification and regression convolutions that produce dense predictions for a pre-defined set of default boxes. We use the same loss function as standard SSD which uses a cross-entropy loss on the classification outputs and a smooth L1 loss on the bounding box regression. The final outputs are refined using non-maxima suppression (NMS) with a threshold of 0.45 and limited to a maximum of the 200 most confident detections.

\subsection{Small objects}\label{small_objects}

As has been noted in other work \cite{lin2017focal}, the performance of SSD degrades when trying to detect smaller objects. To counteract this we adopt an approach similar to that proposed by \cite{muller2018detecting}, duplicating default boxes at sub-divisions of the feature map cells. 

\section{Experimental Setup}

\subsection{Network Training}

We train all of our models from scratch using the ADAM optimiser \cite{kingma2014adam} with \(\beta_{1} = 0.9\), \(\beta_{2} = 0.999\), \(\epsilon = 10^{-8}\) and a learning rate of \(\lambda = 10^{-4}\). We use L2 weight decay of \(10^{-3}\) and train for 50k iterations with a batch size of 16.

We use image augmentation during training to reduce overfitting. Data is flipped left-right with a probability of 0.5 before being cropped with probability 0.5 to between 0.6-1.0 of full image dimensions. If the data is cropped it is resized to full image size before being used. We also randomly modify the hue and saturation of the RGB images. 

To keep feature scales consistent between modalities we scale each modality by its mean and standard deviation calculated over the training set.

\subsection{Evaluation metrics}

We use the average precision metric to measure detection performance with an intersection-over-union (IOU) threshold of 0.5 (higher IOU thresholds are very challenging for instances that are only a few pixels in size). We use the PASCAL VOC2012 \cite{Everingham10} definition for the integration under the curve which sets the precision at recall \(r\) equal to the maximum precision of any recall \(r' \geq r\). In addition, we adopt a similar system to the KITTI benchmark \cite{Geiger2012CVPR} and evaluate performance on separate object size categories. However, unlike KITTI we measure size based on fraction of image area occupied as this applies more equally to objects of different aspect ratios. The size categories used are small (\(<0.25\%\) of image area, approximately \(410px\)), medium (\(0.25\%\)--\(2.5\%\), \(410\)--\(4096px\)) and large (\(>2.5\%\)). Again in contrast to KITTI, there is no minimum size applied for objects as we are interested in detecting very small instances.

\section{Experimental Results}

\subsection{Label quality}

\begin{figure}
\vspace{1em}
\centering
\includegraphics[width=\columnwidth]{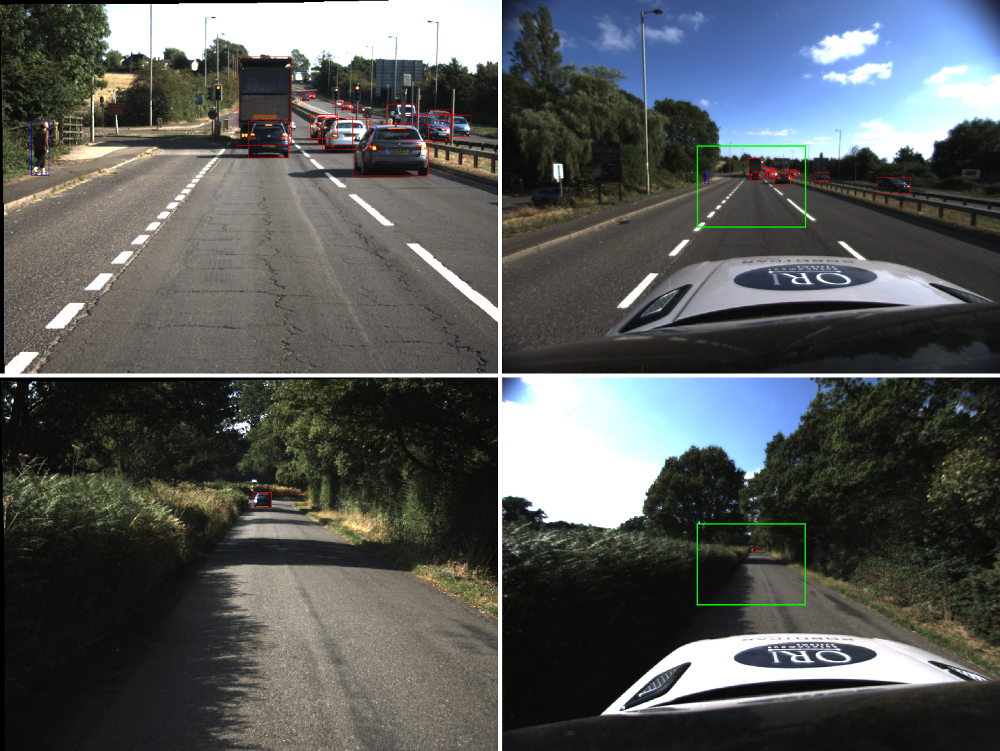} 
\caption{Examples of hand-labelled ground truth using bounding box transfer between two focal lengths. Note the proportionally large number of very small boxes. The green outline in the right-hand images shows the field of view of the left-hand images.}
\label{fig:labelling}
\end{figure}

To assess the quality of the generated labels and the benefits of the multi-camera approach, we compare the generated labels against 161 hand-labelled ground truth frames from the testing set. The hand-labelling interface makes use of the multi-camera bounding box transfer (Sec. \ref{focal_lengths}) to allow the labelling of very small, distant objects (Fig. \ref{fig:labelling}). 

\begin{figure*}[tpb]
    \centering
    \mbox{}\hfill
    \subfloat[\label{fig:label_quality-a}]{\includegraphics[width=43mm]{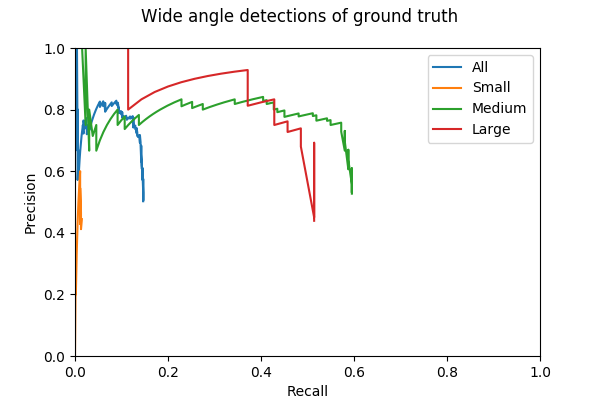}}\hfill
    \subfloat[\label{fig:label_quality-b}]{\includegraphics[width=43mm]{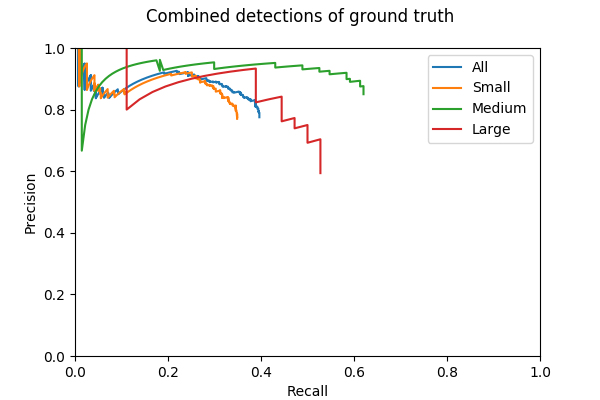}}
    \hfill
    \subfloat[\label{fig:label_quality-c}]{\includegraphics[width=43mm]{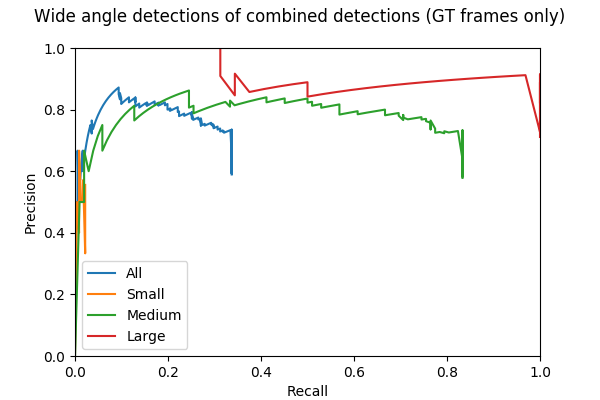}}
    \hfill
    \subfloat[\label{fig:label_quality-d}]{\includegraphics[width=43mm]{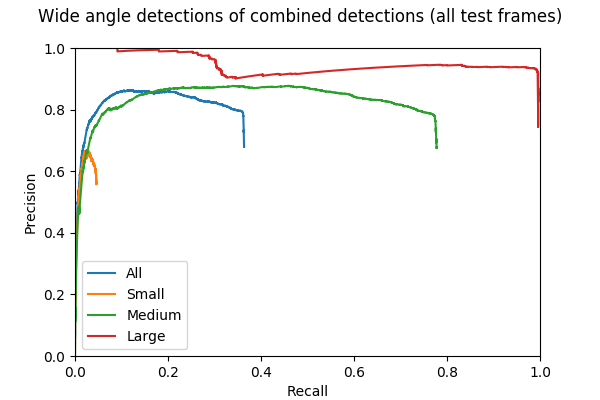}}
    \hfill
    \mbox{}
    \caption{Label quality:
    \protect\subref{fig:label_quality-a} shows the precision-recall of the YOLO object detections from the wide-angle camera only when tested against hand-labelled ground truth images;
    \protect\subref{fig:label_quality-b} shows the precision-recall of YOLO object detections when the detections from the wide-angle and long-focal length cameras are combined;
    \protect\subref{fig:label_quality-c} shows the wide-angle detections tested against the combined set on the same images as those that have been hand-labelled, highlighting which camera contributes which detections --- the wide angle camera contributes the large and the majority of the medium sized objects; and,
    \protect\subref{fig:label_quality-d} expands the test in Fig. \protect\subref{fig:label_quality-c} to cover the whole test set showing that the general pattern is very similar.}%
    \label{fig:label_quality_all}
\end{figure*}

We use the probabilities from the original detector (Sec. \ref{object_detector}) to generate precision-recall curves for four tests. The first two tests check how well our automated labelling recovers the hand-labelled ground truth when using firstly, only wide angle detections (Fig. \ref{fig:label_quality-a}) and secondly, the combined multi-camera detections (Fig. \ref{fig:label_quality-b}). The results (Table \ref{tab:label_quality}) show that there is a significant improvement in the quality of the labels when the combined bounding boxes are used with the improvement particularly marked for small vehicles. 

We then test how each camera contributes labels of different sizes by comparing the detections from the wide angle camera with the combined set of detections. The comparison is initially on only those frames that have hand-labelled ground truth (Fig. \ref{fig:label_quality-c}) and then on the entire test set (Fig. \ref{fig:label_quality-d}). As expected these tests show that the majority of small objects are provided by the narrow field of view camera.

It is worth noting that the precision of the automatically labelled dataset is high (Fig. \ref{fig:label_quality-b}) with the major issue being low recall. While this means that the automatically generated labels cannot be classed as ground truth, from a training perspective \cite{wu2018soft} shows that missing labels do not have as great an effect on performance as might be expected.

\begin{table}\centering
    \caption{Comparisons of labelling methods, average precision (AP).}
    \begin{tabular}{@{}lcccc@{}}\toprule
    \textbf{Comparison} &\textbf{AP}\\ \midrule
        Wide angle detections of ground truth& 0.118\\
        Combined detections of ground truth& 0.360\\
        Wide angle detections of combined detections, GT frames only&0.275\\
        Wide angle detections of combined detections, all test frames&0.308\\
    \bottomrule
    \end{tabular}
    \vspace{-1em}
    \label{tab:label_quality}
\end{table}


\subsection{Trained network performance on generated test set}

\begin{figure}
\vspace{-1em}
\centering
\includegraphics[width=0.60\columnwidth]{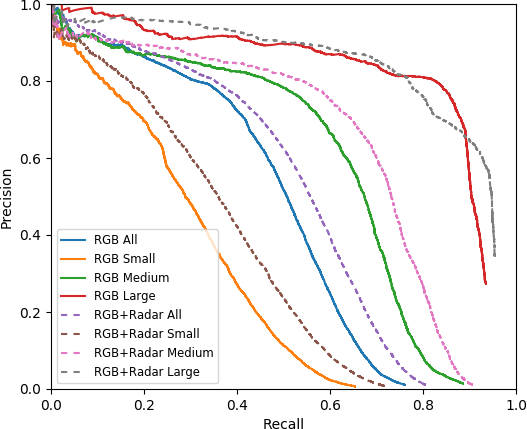} 
\caption{Precision-recall curves for RGB and RGB + Radar (concatenation) networks tested on automatically generated labels. The inclusion of radar produces a significant improvement in the recall of small and medium sized vehicles.}
\label{fig:rgb_vs_rgbradar}
\end{figure}

We test the performance of our network trained with our generated dataset on the test split of the dataset. The results are shown in Table \ref{tab:generated_results}. It can be seen that both radar fusion methods significantly improve the detection of smaller vehicles. The concatenation approach improves performance across all size subsets as shown in Fig. \ref{fig:rgb_vs_rgbradar}.  We believe the performance improvement is due to the radar providing a distinctive signature for moving objects which, in conjunction with the RGB context (the position of the road, for example), allows small objects to be identified. This is in line with the instinctive response when viewing the examples in Fig. \ref{fig:example_data} where the range-rate image gives an excellent indicator of vehicle locations.  

\begin{figure}
\centering
\includegraphics[width=\columnwidth]{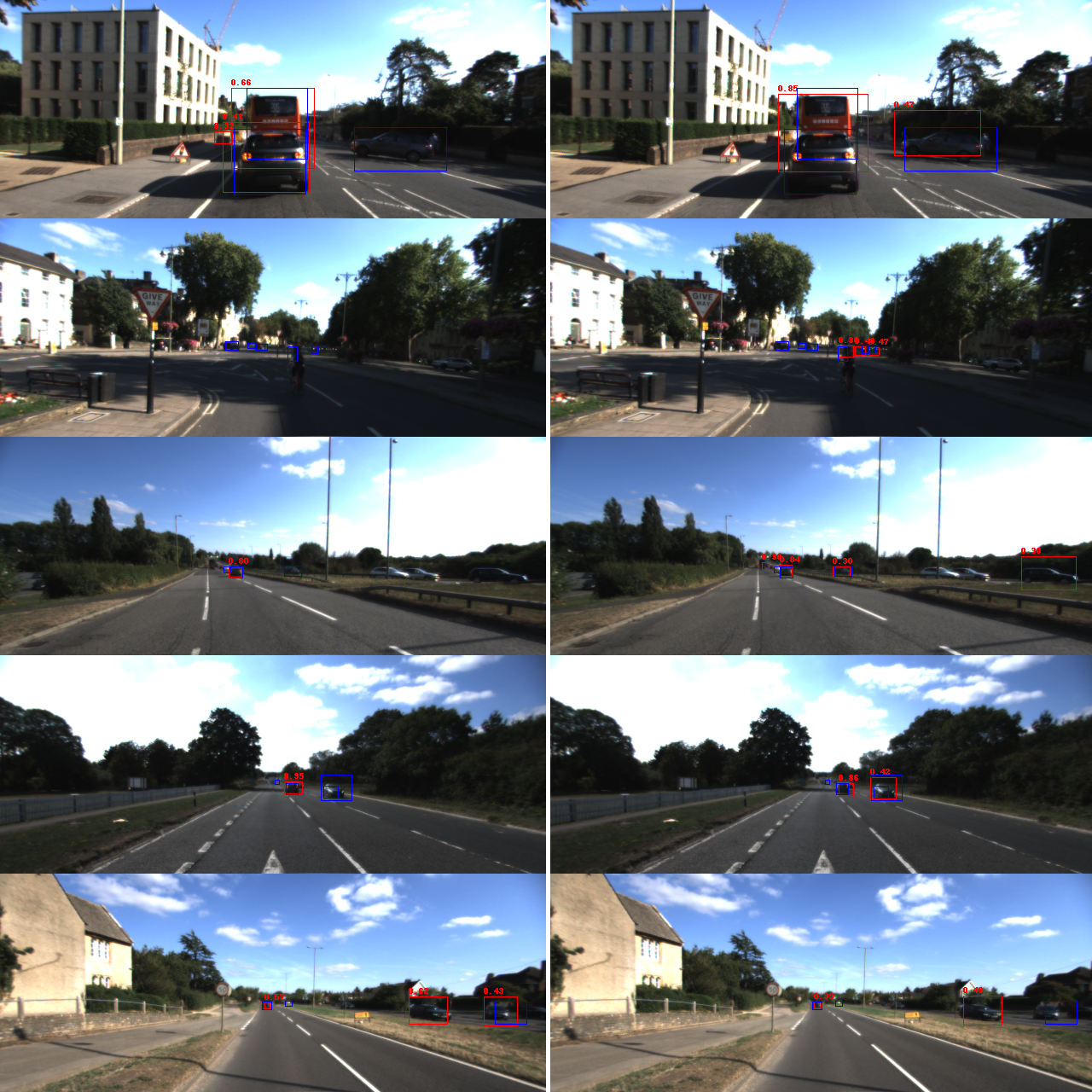} 
\caption{Comparison of detections using both networks. \textit{Left} are detections from the RGB only network. \textit{Right} are from the network using both RGB and radar. Detections are in red with confidences, automatically generated (and hence imperfect) ground truth in blue.}
\label{fig:prediction_comparison}
\vspace{-1em}
\end{figure}

A qualitative comparison of detection performance on a random set of examples is shown in Fig. \ref{fig:prediction_comparison}.

\begin{table}\centering
    \vspace{1em}
    \caption{Performance on generated data by object size, average precision (AP).}
    \begin{tabular}{@{}lcccc@{}}\toprule
    \textbf{Data type} & \textbf{Small} & \textbf{Medium} & \textbf{Large} & \textbf{All}\\ \midrule
        RGB only & 0.287 & 0.588 & 0.816 & 0.460\\
        RGB + Radar, concatenation & \textbf{0.346} & \textbf{0.644} & \textbf{0.833} & \textbf{0.506}\\
        RGB + Radar, element-wise & 0.327 & 0.599 & 0.723 & 0.461\\
    \bottomrule
    \end{tabular}
    \label{tab:generated_results}
\end{table}

\subsection{Trained network performance on hand-labelled test set}

\begin{figure}
\vspace{-1em}
\centering
\includegraphics[width=0.60\columnwidth]{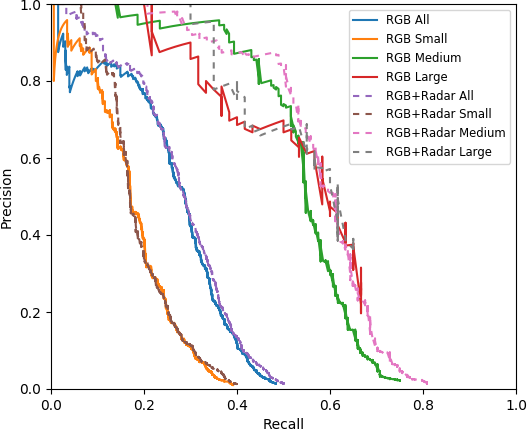} 
\caption{Precision-recall curves for RGB and RGB + Radar (element-wise) networks tested on hand-labelled data. The lack of smoothness in the curves is due to the small number of examples in the ground-truth set.}
\label{fig:rgb_vs_rgbradar_hand}
\end{figure}

\begin{figure}
\centering
\includegraphics[width=0.67\columnwidth]{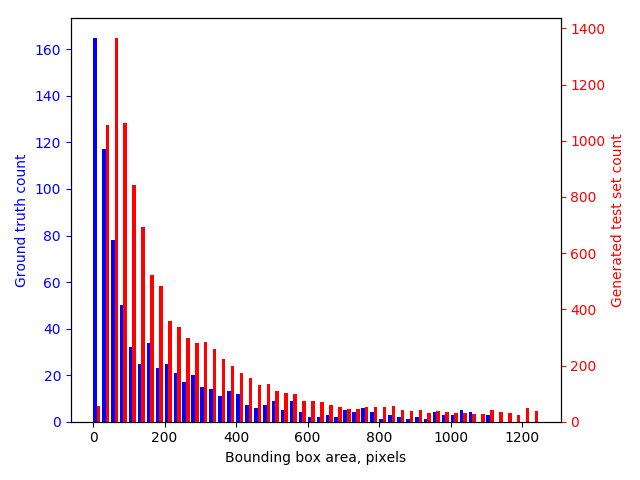} 
\caption{Comparison of the size of label bounding boxes between the hand-labelled ground truth (\textit{blue}) and the generated test set (\textit{red}). The ground truth set has proportionally more extremely small objects. The histogram is truncated so only the small end is shown.}
\label{fig:joint_sizes}
\end{figure}

We next test on the small subset of hand-labelled data. Table \ref{tab:hand_results} shows that both radar fusion methods provide an increase in performance, with the element-wise version outperforming the concatenation method. Fig. \ref{fig:rgb_vs_rgbradar_hand} compares the RGB and element-wise fusion networks. In general the results are lower than on the generated data. This may be partially due to the higher proportion of very small objects in the ground truth set (Fig. \ref{fig:joint_sizes}).

All three networks significantly outperform the original detector which had an average precision of only 0.118 (Table \ref{tab:label_quality}), showing the benefit of the multi-camera labelling process.

\begin{table}\centering
    \caption{Performance on hand-labelled subset by object size, average precision (AP).}
    \begin{tabular}{@{}lcccc@{}}\toprule
    \textbf{Data type} & \textbf{Small} & \textbf{Medium} & \textbf{Large} & \textbf{All}\\ \midrule
        RGB only & 0.178 & 0.541 & 0.526 & 0.263\\
        RGB + Radar, concatenation & 0.170 & 0.558 & 0.542 & 0.265\\
        RGB + Radar, element-wise & \textbf{0.188} & \textbf{0.577} & \textbf{0.544} & \textbf{0.279} \\
    \bottomrule
    \end{tabular}
    \vspace{-1em}
    \label{tab:hand_results}
\end{table}

\section{Conclusions and further work}

We have introduced a process for automatically labelling a new dataset by combining detections from multiple cameras. We have also demonstrated how that can be used to train a network which fuses radar scans with camera images to improve detection performance. In addition we have shown that by learning from a combined set of detections we are able to exceed the performance of the original detector.

In future work there are a number of avenues that we intend to explore. First, the image-like radar representation used in this work is very simple and there may be other representations that might be better suited to the sparsity of the radar data. Secondly, there may be benefits to using consecutive frames both to filter noisy labels and to reduce the impact of radar noise. Finally, given that our labels are automatically generated, they will contain some noise. While \cite{rolnick2017deep} showed that, given enough examples, deep learning is sufficiently robust to learn an accurate model despite large amounts of label noise, there have been a number of methods proposed for handling such noise (e.g. \cite{jiang2017mentornet}) that would be worth exploring.


\section*{Acknowledgements}
The authors would like to thank Martin Engelcke and colleagues for the object detector implementation used to generate the dataset. We gratefully acknowledge the JADE-HPC facility for providing the GPUs used in this work. Paul Newman is funded by the EPSRC Programme Grant EP/M019918/1.

\newpage

\bibliographystyle{IEEEtran}
\input{bibliography.bbl}


\end{document}